\documentclass[letterpaper, 10 pt, conference]{ieeeconf}  

\IEEEoverridecommandlockouts                              

\usepackage{amsmath} 
\usepackage{cite} 
\usepackage{graphicx}
\usepackage{amsmath,amsfonts,amssymb}
\usepackage{bm}
\usepackage{url}
\usepackage{breakurl}
\usepackage[breaklinks]{hyperref}
\usepackage{float}
\usepackage{stfloats} 
\usepackage{makecell}
\usepackage[font=small,skip=2pt]{caption}

\title{\LARGE \bf
PHANTOM Hand
}

\author{
Teng Yan$^{1}$,
Jiongxu Chen$^{1}$,
Qixiang Hua$^{1}$,
Yue Yu$^{1}$,
Zihang Wang$^{1}$,
Yaohua Liu$^{1}$,
and Bingzhuo Zhong$^{1}$\thanks{Corresponding author.}%
\\
$^{1}$The Hong Kong University of Science and Technology (Guangzhou)
\\
{\tt\small tyan497@connect.hkust-gz.edu.cn, bingzhuoz@hkust-gz.edu.cn
}
}

\begin{document}

\maketitle
\thispagestyle{empty}
\pagestyle{empty}

\begin{abstract}
Tendon-driven underactuated hands excel in adaptive grasping but often suffer from kinematic unpredictability and highly non-linear force transmission. This ambiguity limits their ability to perform precise free-motion shaping and deliver reliable payloads for complex manipulation tasks. To address this, we introduce the PHANTOM Hand (Hybrid Precision-Augmented Compliance): a modular, 1:1 human-scale system featuring 6 actuators and 15 degrees of freedom (DoFs).

We propose a unified framework that bridges the gap between precise analytic shaping and robust compliant grasping. By deriving a sparse mapping from physical geometry and integrating a mechanics-based compensation model, we effectively suppress kinematic drift caused by spring counter-tension and tendon elasticity. This approach achieves sub-degree kinematic reproducibility for free-motion planning while retaining the inherent mechanical compliance required for stable physical interaction.

Experimental validation confirms the system's capabilities through (1) kinematic analysis verifying sub-degree global accuracy across the workspace; (2) static expressibility tests demonstrating complex hand gestures; (3) diverse grasping experiments covering power, precision, and tool-use categories; and (4) quantitative fingertip force characterization. The results demonstrate that the PHANTOM hand successfully combines analytic kinematic precision with continuous, predictable force output, significantly expanding the payload and dexterity of underactuated hands. To drive the development of the underactuated manipulation ecosystem, all hardware designs and control scripts are fully open-sourced for community engagement.
\end{abstract}

\section{Introduction}
In domains such as mobile manipulation, human–robot collaboration, and embodied intelligence, the capability of an end-effector to reproducibly achieve target joint configurations under limited actuation fundamentally constrains the upper bound of system performance. Fully-actuated dexterous hands offer ample controllability and bandwidth through one-to-one actuation, but their mass, volume, power consumption, and wiring complexity escalate with the number of DoFs, restricting their deployment on mobile platforms and in low-cost applications. In contrast, underactuated tendon-driven systems provide a lightweight, safe, and cost-effective solution through structural synergy and inherent compliance. 

However, while this mechanical compliance excels during the contact phase of grasping, achieving a precise and reliable mapping from low-dimensional actuation to high-dimensional configuration has long been impeded by interrelated real-world factors. First, tendon routing and wrap angles vary with posture, causing the equivalent radius to depend strongly on configuration and rendering simple proportional mappings invalid. Second, the inevitable friction and hysteresis in the rope-pulley-sheath system induce path dependence and bidirectional bias. Third, the micro-elasticity of the tendons and structure introduces an additional deformation pathway, creating a systematic discrepancy between analytical geometric models and physical behavior. A lack of systematic compensation for these effects results in an expression gap: the theoretically reachable configurations do not match the physically reproducible configurations, thereby hindering precise free-motion shaping and degrading the robust force transmission required for tool-use affordances.

\begin{figure}[t]
    \centering
    \includegraphics[width=1.0\columnwidth]{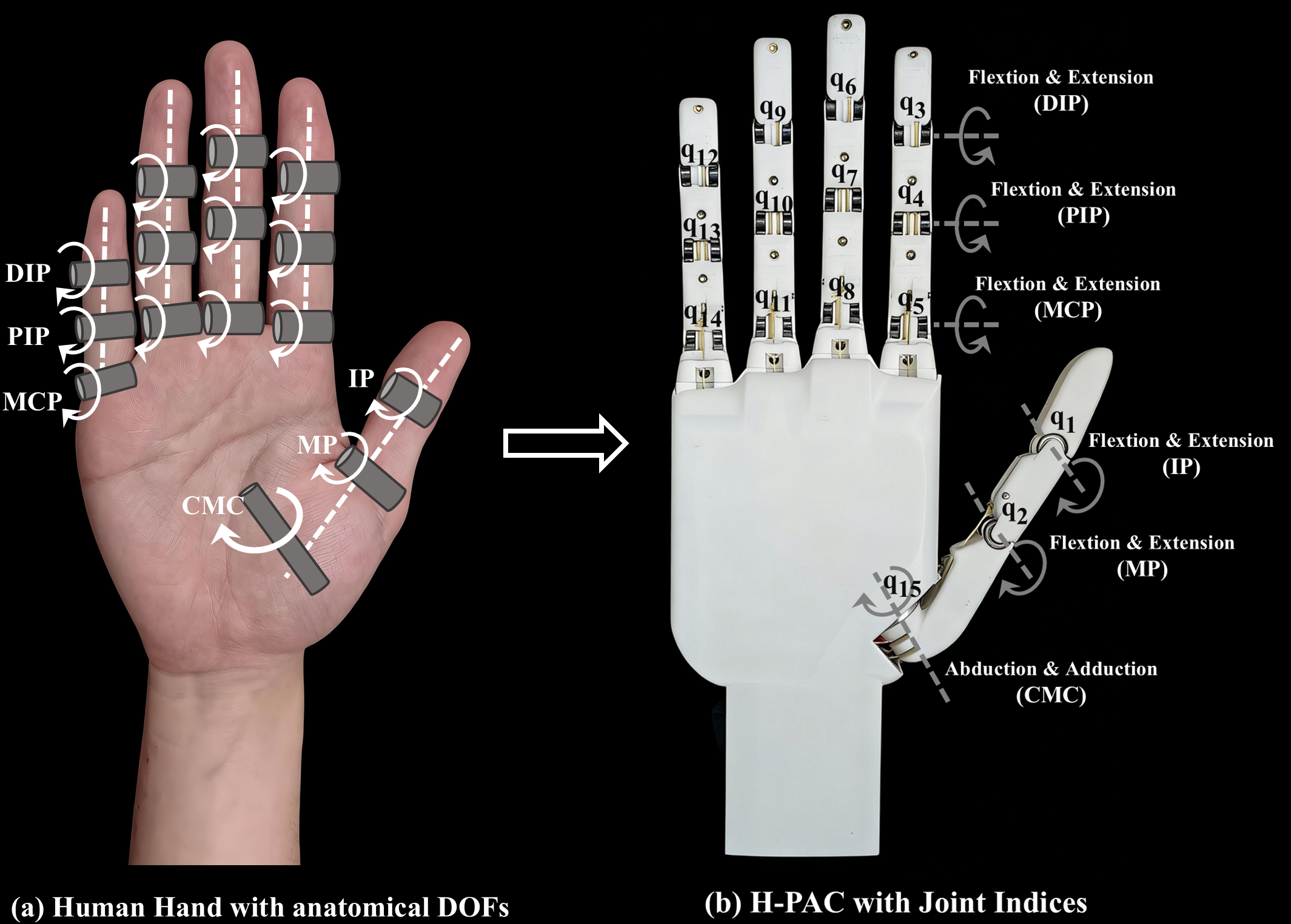}
    \caption{\textbf{Overview and DOF mapping of PHANTOM Hand.}  
    (a) Human hand annotated with anatomical DOFs (MCP/PIP/DIP, thumb MP/IP, and CMC).  
    (b) The PHANTOM hand with 15 joint DOFs $q_{1}\!-\!q_{15}$: four long digits each with MCP/PIP/DIP flexion–extension, and the thumb with IP/MP flexion–extension and CMC abduction–adduction.}
    \label{fig:phantom_overview}
\end{figure}

To bridge this gap, this paper introduces the \textbf{PHANTOM Hand (Hybrid Precision-Augmented Compliance)}, a 6-actuator, 15-DoF, 1:1 human-scale, magnetically modular tendon-driven dexterous hand (Fig.~\ref{fig:phantom_overview}). The four long fingers employ a figure-eight tendon routing, enabling coordinated flexion across the MCP, PIP, and DIP joints. The thumb uses two actuators: one tendon-driven actuator for IP/MP flexion, and another directly actuated joint where the servo output axis is collocated with the CMC joint axis for CMC opposition/abduction-adduction. Based on this hardware topology, this work constructs an invertible block-sparse analytical mapping from 6 actuators to 15 degrees of freedom. 
To systematically suppress the full-range deviations arising from tendon micro-elasticity and spring counter-tension, a mechanics-based kinematic compensation model is integrated. Without adding tension or force sensors or modifying the hardware, this pipeline achieves sub-degree, reproducible configuration expression while maintaining real-time control.

Based on this design and methodology, the expressiveness of the configuration and the performance of the task are evaluated on static hand gestures and a comprehensive set of grasps that encompasses the power, precision and tool-specific categories. Furthermore, quantitative fingertip force characterization is employed to demonstrate the system's robust payload capacity. Sweep-curve analysis is also utilized to assess the full-range consistency of the analytic plus mechanics-based compensation approach, confirming the hand's hybrid advantage in both precision and power. The complete CAD models, ROS2 files, and calibration scripts will be open-sourced to facilitate community collaboration and foster a thriving research ecosystem.

Key contributions of this work include:
\begin{itemize}
    \item \textbf{Interpretable and invertible analytical mapping:} A mapping from 6 actuators to 15 joints, derived from tendon routing geometry, which enables closed-form forward and inverse kinematics.  
    \item \textbf{Mechanics-based compliance compensation:} A physical model accounting for spring counter-tension and tendon elasticity to systematically suppress full-range kinematic deviations without additional sensing, achieving sub-degree accuracy.  
    \item \textbf{Human-scale, magnetically modular platform:} A 1:1 human-scale hand with fingers magnetically attached to the palm, enabling rapid assembly, maintenance, and reconfiguration. 
    \item \textbf{Validation of hybrid capabilities:} Comprehensive experimental validation demonstrating that the PHANTOM hand successfully combines analytic kinematic precision with continuous, robust force output, expanding the payload and dexterity of underactuated hands.
\end{itemize}

\section{Related Work}
\subsection{Motivation for Compact and Underactuated Designs}
Fully actuated dexterous hands assign an independent actuator to each degree of freedom, providing high expressivity and control bandwidth, as exemplified by the Gifu Hand series\cite{Mouri2011, Kawasaki2002} and the DLR Hand Arm System \cite{Grebenstein2012}. However, such designs incur significant burdens in terms of weight, volume, power consumption, and cost \cite{Grebenstein2012, Tokunaga2025}. To balance performance and feasibility, recent research has increasingly focused on compact and underactuated topologies \cite{Piazza2019, Brown2007}, where multiple joints are coordinated by a small number of actuators through mechanical coupling, tendon routing, or elastic elements \cite{Kim2021, Schunk2025, Bridgwater2012, Wall2017}, thereby enhancing deployability and safety. Compared to fully actuated hands, underactuated approaches demonstrate advantages in grasp robustness and compliant interaction \cite{He2019}, but impose new challenges in modeling and calibration to achieve precise joint control.

\subsection{Mapping and Calibration: From Empirical Linear Models to Analytical Approaches}
Early research efforts commonly employed empirical linear scaling or synergistic models to establish the mapping from actuator strokes to finger joint angles \cite{Telegenov2015}. These approaches often use low-dimensional synergy projections of human hand motions \cite{Dwivedi2020, Romero2013} to reduce control dimensionality, thus facilitating rapid, although approximate, pose representations \cite{Santos2025, Liu2008}. An alternative line of research has focused on their sensitivity to manufacturing tolerances and assembly variations, which substantially restricts cross-prototype transferability. A parallel line of inquiry has focused on explicit closed-form models of the tendon length-joint-angle relationship, derived from geometry and equivalent radius formulations. Although this approach provides interpretable and reusable actuator–joint mappings\cite{kim2020joint, sainul2016three}, practical tendon routing introduces non-idealities, such as guiding, wrapping, and slack, which, if left uncalibrated, lead to systematic drift of analytical models, particularly across large joint ranges.

\subsection{Sensing and Data-Driven Compensation}
High-fidelity calibration is often achieved with tension/displacement sensors or optical measurement systems to jointly identify elastic deformation, frictional effects, and hysteresis in the actuation system \cite{Liu2008, Johansson2009, Dargahi2004}. End-to-end learning has also been applied to directly model the mappings between actuators and joints or end-effectors \cite{Pinto2016, Kalashnikov2018}. While the former ensures physical interpretability, it entails high hardware complexity and limited robustness, undermining practicality in real-world applications. The latter offers strong task adaptability but demands large-scale data and computation \cite{Mohammed2020}, and lacks interpretability and cross-prototype transferability. In contrast, augmenting analytical mappings with low-order, small-sample angle-dependent bias terms provides a practical trade-off between transparency and usability, making it particularly well-suited for underactuated hand engineering.

\begin{figure}[h!]
    \centering
    \includegraphics[width=1.0\columnwidth]{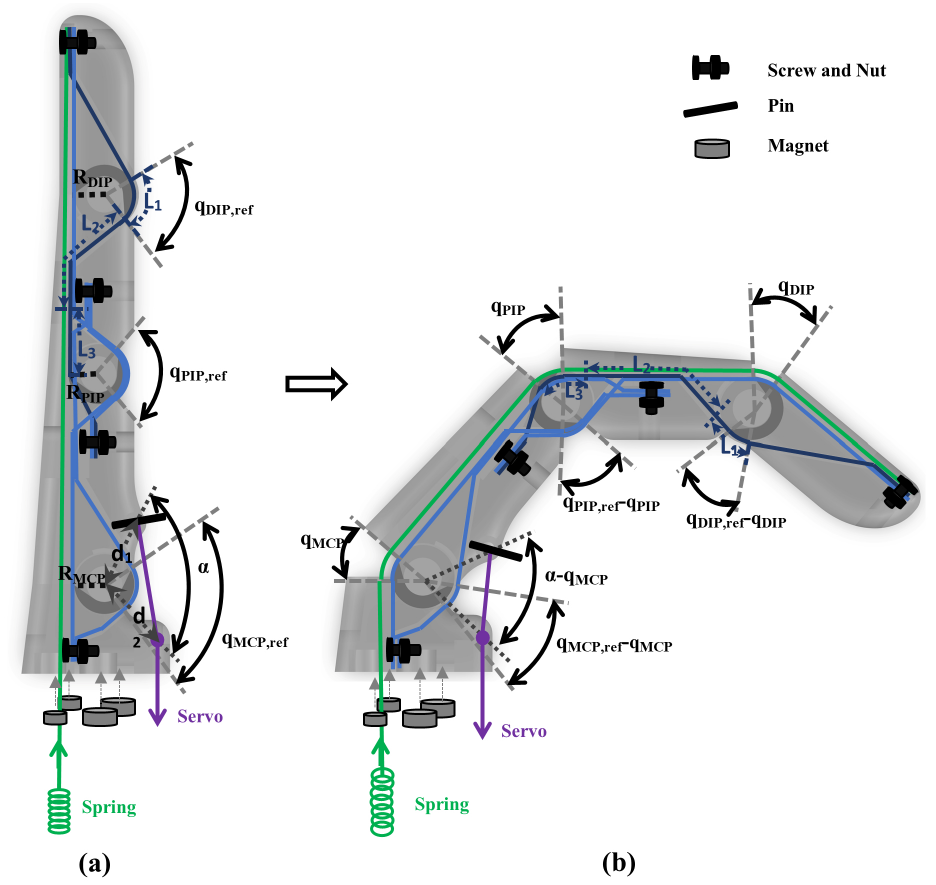}
    \caption{\textbf{Tendon routing and analytic modeling for four fingers.}
    Four-finger figure-eight routing from full extension to an intermediate flexion: straight segments $L_k$, joint pulleys of radius $R_i$, cables connected to the servo hubs (purple), the passive "cable-spring" return path (green), parameters related to the closed-form geometric expression of the MCP joint ($\alpha$, $d_{1}$, $d_{2}$), and joint angles $q_{\text{MCP}}$, $q_{\text{PIP}}$, $q_{\text{DIP}}$.  $q_{i,\text{ref}}$ (where $i=\{\text{DIP},\text{PIP},\text{MCP}\}$) denotes the structural geometric limits of the independent finger module for analytical derivation, while the actual operational range limits are constrained by the palm housing to prevent excessive flexion.}
    \label{fig:fingers}
\end{figure}

The contribution of PHANTOM Hand lies in proposing a unified framework that integrates analytical geometric mapping, single-sweep calibration, and angular compensation, enabling zero additional sensing, sub-degree pose reproducibility, and task generalization on a 6-actuator/15-DoF hardware platform—ultimately delivering an underactuated yet precisely expressive biomimetic dexterous hand.

\section{methodology}
\subsection{Hand design and modeling}
\begin{figure}[h!]
    \centering
    \includegraphics[width=1.0\columnwidth]{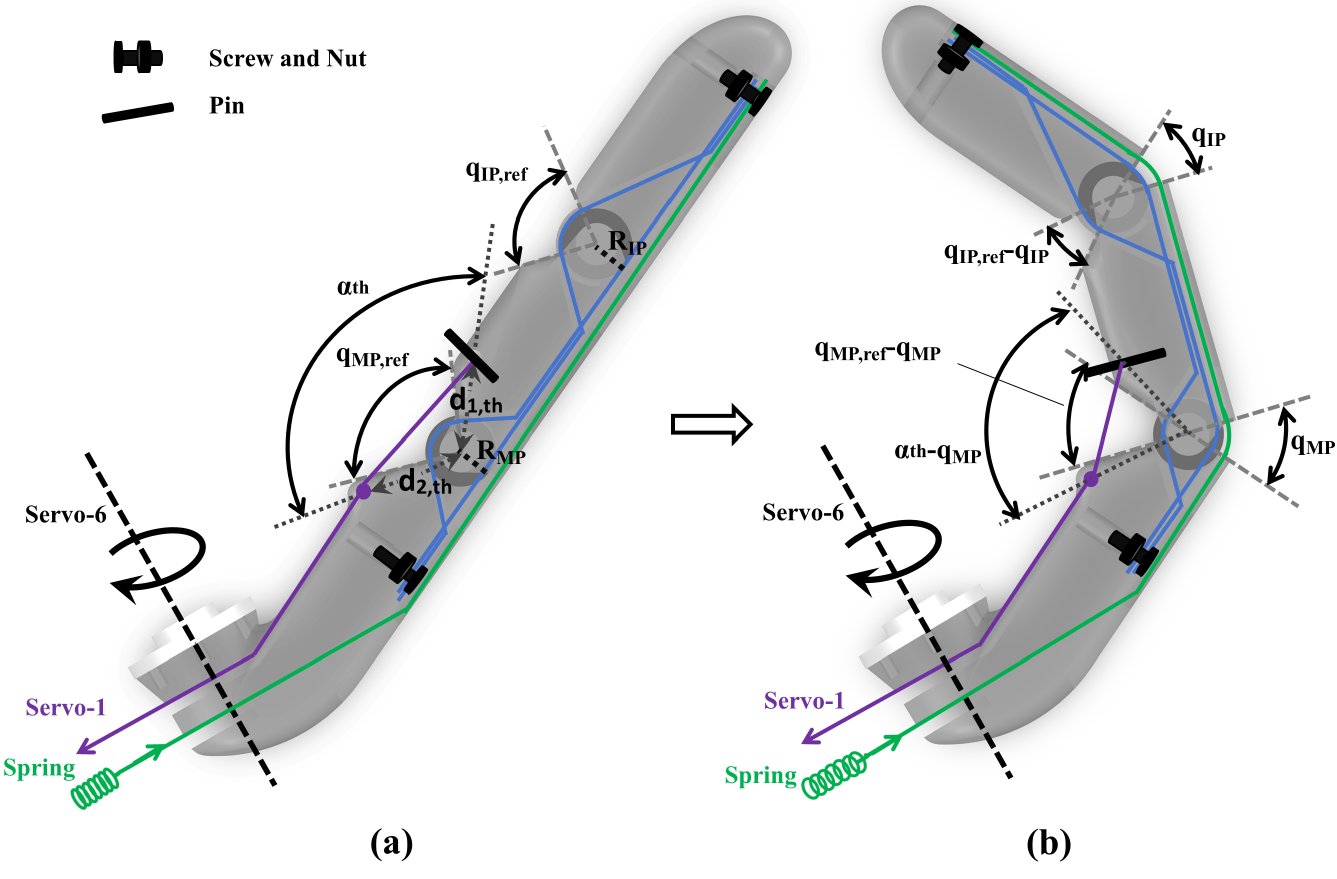}
    \caption{\textbf{Tendon routing and analytic modeling for thumb.}
    Thumb figure-eight routing from full extension to an intermediate flexion and thumb dual-actuation: 1) Servo-1 drives MP/IP with figure-eight routing (parameters $\alpha_{\mathrm{th}}$, $d_{1,\mathrm{th}}$, $d_{2,\mathrm{th}}$). The model includes joint pulleys of radius $R_i$, cables connected to the servo hubs (purple), the passive "cable-spring" return path (green), and joint angles $q_{\text{MP}}$, $q_{\text{IP}}$. $q_{i,\text{ref}}$ (where $i=\{\text{DP},\text{MP}\}$) denotes the structural geometric limits of the independent finger module for analytical derivation, while the actual operational range limits are constrained by the palm housing to prevent excessive flexion. 2) Servo-6 drives the CMC joint, with its output axis coaxial to the CMC joint axis and the servo horn directly rigidly attached to the CMC joint.}
    \label{fig:thumb}
\end{figure}

The PHANTOM hand is designed with a 6-actuator, 15-DoF tendon-driven topology. Four fingers (index, middle, ring, and pinky) are individually actuated by servos 2–5. The tendon for each of these fingers is routed in a figure-eight configuration across the distal interphalangeal (DIP), proximal interphalangeal (PIP), and metacarpophalangeal (MCP) joints, as depicted in Fig.~\ref{fig:fingers}, enabling their coupled flexion. The thumb's interphalangeal (IP) and metacarpophalangeal (MP) joints are flexed via a tendon-driven actuation by servo 1. For its part, the carpometacarpal (CMC) joint's transverse abduction/adduction is actuated directly by servo 6, also shown in Fig.~\ref{fig:thumb}.

Additional hardware features include:
\begin{itemize}
\item \textbf{Passive Cable–Spring Return Mechanism:} Each finger is equipped with a passive "cable-spring" return path. Upon release of tendon tension, the spring passively returns the finger segments to their fully extended position.
\item \textbf{Magnetic Modularity:} Four fingers attach to the palm with magnetic quick-release connections, which facilitate maintenance and enable replacement with fingers of varying sizes.
\item \textbf{Low-Friction Biomimetic Design:} Finger bone lengths match those of the human hand (1:1 scale). Adjacent fingers are arranged with a 5° angular offset to mimic the natural arc of the human palm. The metacarpal heads have a stepped coronal plane, and bearings are used at each joint to reduce friction.
\item \textbf{Human-Like Joint Ratio Design:}
According to human finger characteristics, the radius ratio of DIP:PIP:MCP joints for the four fingers is set to 5:4:5, achieving an ideal rotation angle ratio of 4:5:4. For the thumb, the IP:MP joint radius ratio is 5:4, yielding a 4:5 angle ratio, which are consistent with the kinematic behavior of the human hand. Details are provided in \ref{Four-Finger Model} and \ref{Thumb Model}.
\end{itemize}

For subsequent analysis and control, the following notation is adopted:

\begin{itemize}
    \item \textbf{$\mathbf{u}=[u_{1}, \dots, u_{6}]^\top \in \mathbb{R}^{6}$}: servo angles (rad); 
    $\mathbf{r} = [r_{1}, \dots, r_{6}]^\top$ are the corresponding pulley radii (m).  

    \item $\mathbf{q} = [q_{1}, \dots, q_{15}]^\top \in \mathbb{R}^{15}$: joint angles (rad), indexed as in Fig.~\ref{fig:phantom_overview};
    $\mathbf{R} = [R_1, \dots, R_{15}]^\top$: joint radii (m), with the CMC joint excluded from the 15-joint set, which govern the tendon length variation via the figure-eight routing and guiding geometry shown in Fig.~\ref{fig:fingers} and Fig.~\ref{fig:thumb}.

%

    \item \textbf{$\mathbf{H} \in \mathbb{R}^{15 \times 6}$}: primary mapping matrix (block-sparse, with only 15 non-zero entries; each non-zero entry is a function of a single $u_j$). 

    \item \textbf{$\mathbf{b(q)} = [b_1(q_1), \dots, b_{15}(q_{15})]^\top \in \mathbb{R}^{15}$}: compensation term (rad), defined as the difference between the ideal joint angle $\mathbf{q_{\text{ideal}}}$ and the compensated joint angle $\mathbf{q_{\text{comp}}}$, i.e., $\mathbf{b(q)} = \mathbf{q_{\text{ideal}}} - \mathbf{q_{\text{comp}}}$; the compensation term is only non-zero for interphalangeal joints (IP, PIP, DIP) excluding MCP and thumb MP joints.

    \item \textbf{$k_s$ (N/m)}: stiffness of the restoring spring.
    \item \textbf{$EA$ (N)}: axial stiffness of the tendon, where $E$ is the Young's modulus of the tendon material and $A$ is the cross-sectional area.
\end{itemize}

\subsection{Single-Finger Model}
\subsubsection{Four-Finger Model}\

\label{Four-Finger Model}
\textbf{Tendon length conservation under the figure-eight routing.} 
Due to the figure-eight routing of the tendons between adjacent joints, the extension and contraction of the tendon segments at neighboring joints mutually compensate during flexion. According to the geometric constraints, the incremental tendon lengths at each joint satisfy $R_{\mathrm{DIP}} \Delta q_{\mathrm{DIP}} = R_{\mathrm{PIP}} \Delta q_{\mathrm{PIP}}$ and $R_{\mathrm{PIP}} \Delta q_{\mathrm{PIP}} = R_{\mathrm{MCP}} \Delta q_{\mathrm{MCP}}$. Consequently, in an ideal state neglecting initial tension, the joint angles follow the coupled relationship
\[
R_{\mathrm{DIP}}\,q_{\mathrm{DIP}}
=R_{\mathrm{PIP}}\,q_{\mathrm{PIP}}
=R_{\mathrm{MCP}}\,q_{\mathrm{MCP}}.
   \tag{2}\]
By setting $R_{\mathrm{DIP}} : R_{\mathrm{PIP}} : R_{\mathrm{MCP}} = 5:4:5$ , the joints exhibit an equal-angle synergy: 
\[
{\;q_{\mathrm{DIP}}:q_{\mathrm{PIP}}:q_{\mathrm{MCP}}=4:5:4 .\;}  \tag{3}
\]

\textbf{Closed-form expression of the MCP joint.}
As shown in Fig.~\ref{fig:fingers}(a), (b), let $d_1,d_2$ denote the fixed distances from the guides to the MCP rotation center, and let $\alpha$ be the initial included angle. The initial chord length is  
\[
L_0=\sqrt{d_1^2+d_2^2-2d_1d_2\cos\alpha}. \tag{4}
\]
The tendon retraction by servo $j$ is $\Delta L=r_j u_j$, giving current chord length $L=L_0-\Delta L$.  
By the cosine law, the instantaneous angle is given by
\[
\varphi=\arccos\!\left(\frac{d_1^2+d_2^2-L^2}{2d_1 d_2}\right). \tag{5}
\]
Flexion is defined as a decrease of the included angle from its initial value $\alpha$, the MCP rotation is  
\[
{\;q_{\mathrm{MCP}}=\alpha-\arccos\!\left(\frac{d_1^2+d_2^2-(L_0-r_j u_j)^2}{2d_1d_2}\right).\;}
 \tag{6}\]

\subsubsection{Thumb Model}\

\label{Thumb Model}
\textbf{Coupled IP–MP Flexion.} 
Fig.~\ref{fig:thumb} illustrates the thumb moving from extension to an intermediate flextion. The two tendons at IP/MP joints are also routed in a figure-eight configuration, with a design ensuring $R_{\mathrm{IP}}:R_{\mathrm{MP}}=5:4$, thus  
\[
{\;q_{\mathrm{IP}}:q_{\mathrm{MP}}=4:5 .\;}
 \tag{7}\]
Let the thumb guiding parameters be $d_{1,\mathrm{th}}, d_{2,\mathrm{th}}, \alpha_{\mathrm{th}}$, as in Fig.~\ref{fig:thumb}(a). Analogous to the finger MCP case, applying the cosine law yields,  
\begin{align*}
q_{\mathrm{MP}} &= \alpha_{\mathrm{th}}
 - \arccos\!\left(
   \frac{d_{1,\mathrm{th}}^2+d_{2,\mathrm{th}}^2-(L_{0,\mathrm{th}}-r_1u_1)^2}
        {2d_{1,\mathrm{th}}d_{2,\mathrm{th}}}
   \right). \tag{8}
\end{align*}

\textbf{CMC abduction.}
The abduction of the CMC joint is implemented via a direct drive mechanism. The proximal end of the thumb's PLP (Proximal Phalanx) is coupled directly to the servo output axis. The relationship between the joint angle $q_{CMC}$ and the servo input $u_6$ is expressed as
\[
{\;q_{\mathrm{CMC}}=g_{\mathrm{CMC}}\,u_6,\quad g_{\mathrm{CMC}}=1.\;}  \tag{9}
\]

\subsection{Sparse Mapping Model}
By aggregating the relations of four fingers and the thumb, a \(15 \times 6\) function-type mapping matrix $\mathbf{H}(\mathbf{u})$ is obtained:

\begin{itemize}
    \item Four-finger block (each $3 \times 1$): the three rows corresponding to DIP/PIP/MCP of the same finger share the same functional term 
  $\mathbf{H}_{ij}(u_j) = \tfrac{q_{\mathrm{finger}}}{u_j}$; 

  \item Thumb flexion block ($2 \times 1$): $\mathbf{H}_{i1}(u_1) = \tfrac{q_{\mathrm{thumb}}}{u_1}$;

  \item CMC block ($1 \times 1$): $\mathbf{H}_{15,6} = g_{\mathrm{CMC}} = 1$ (constant).  

\end{itemize}

Thus, the preliminary mapping relation is expressed as
\[
\mathbf{q_{ideal}} = \mathbf{H}(\mathbf{u})\,\mathbf{u}. \tag{10}
\]

\subsection{Mechanics-Based Compensation Model}
\subsubsection{Four-Finger Model}\

While the joint angles theoretically follow a $4:5:4$ coupling ratio, the introduction of a fingertip restoring spring for self-extension significantly impacts the transmission accuracy in practical applications. As the finger curls, the spring tension $T_{spr}$ acts upon the entire transmission chain, inducing non-negligible elastic strain in the driving and coupling cables. This strain accumulates along the transmission path, causing the actual joint angles to deviate from the ideal values, with more pronounced discrepancies observed in the distal joints.

According to Hooke's Law, the restoring force generated by the spring is proportional to the total excursion:
\[T_{spr} = k_s (R_{\mathrm{DIP}}\,q_{\mathrm{DIP}}
+ R_{\mathrm{PIP}}\,q_{\mathrm{PIP}}
+ R_{\mathrm{MCP}}\,q_{\mathrm{MCP}}). \tag{11}\]




The elastic elongation $\delta$ of the cable for each segment follows the fundamental equation $\delta = \frac{T L}{EA}$. Due to this deformation, the displacement intended for the driven joints is partially absorbed. Since the MCP joint is directly driven by the actuator, the extremely small deformation of the driving cable can be neglected, leading to $q_{comp,\mathrm{MCP}} \approx q_{ideal,\mathrm{MCP}}$. For the PIP and DIP joints, the angular deviation $b(q_j)$ between their compensated actual angles $q_{comp,j}$ and ideal angles $q_{ideal,j}$ is defined as
 \[b(q_j) = \frac{\delta_j}{R_j} = \frac{T_{spr} \cdot \sum L_{path}}{EA \cdot R_j},  \tag{12} \]
where $ j=\{\text{PIP},\text{DIP}\} $, and the actual joint angle is given by $q_{comp,j}=q_{ideal,j} - b(q_j)$.

Integrating the aforementioned constraints, the system is modeled as a set of quasi-static equilibrium equations. By incorporating the elastic deviation terms, the following matrix equation is established:
\[\mathbf{M} \mathbf{q_{comp}} = \mathbf{A} q_{comp,\mathrm{MCP}}.   \tag{13}\]

where $\mathbf{q} = [q_{comp,\mathrm{MCP}}, q_{comp,\mathrm{PIP}}, q_{comp,\mathrm{DIP}}]^T$ is the actual joint angle vector, and $\mathbf{A} = [1, 1.25, 1]^T$ is the ideal coupling vector determined by the hardware geometry. The system coupling matrix $\mathbf{M}$ is defined as

$$\mathbf{M} = \begin{bmatrix} 1 & 0 & 0 \\ C_\mathrm{PIP} R_\mathrm{MCP} & 1 + C_\mathrm{PIP} R_\mathrm{PIP} & C_\mathrm{PIP} R_\mathrm{DIP} \\ C_\mathrm{DIP} R_\mathrm{MCP} & C_\mathrm{DIP} R_\mathrm{PIP} & 1 + C_\mathrm{DIP} R_\mathrm{DIP} \end{bmatrix},$$
where $C_\mathrm{PIP} = \frac{k_s L_{\mathrm{MCP},\mathrm{PIP}}}{EA R_\mathrm{PIP}}$ and $C_\mathrm{DIP} = \frac{k_s (L_{\mathrm{MCP},\mathrm{PIP}} + L_{\mathrm{PIP},\mathrm{DIP}})}{EA R_\mathrm{DIP}}$ represent the cumulative structural compliance (Let $L_{\mathrm{MCP},\mathrm{PIP}}$ and $L_{\mathrm{PIP},\mathrm{DIP}}$ be the path lengths between MCP-PIP and PIP-DIP joints, respectively). By inverting $\mathbf{M}$, the precise mapping from the servo input to the actual joint responses is obtained as $\mathbf{q_{comp}} = \mathbf{M}^{-1} \mathbf{A} q_{comp,\mathrm{MCP}}$, enabling physical compensation for flexible transmission errors.

\subsubsection{Thumb Model}\

Since the flexion and extension of the thumb involve only two active joints ($q_\mathrm{MP}$ and $q_\mathrm{IP}$), its mechanics-based compensation model can be formulated as a second-order reduced version of the generalized model described above.

the system coupling matrix $\mathbf{M}$ is simplified into a $2 \times 2$ form. The matrix equation is expressed as

\[\begin{bmatrix} 1 & 0 \\ C R_\mathrm{MP} & 1 + C R_\mathrm{IP} \end{bmatrix} \begin{bmatrix} q_{comp,\mathrm{MP}} \\ q_{comp,\mathrm{IP}} \end{bmatrix} = \begin{bmatrix} 1 \\ 1.25 \end{bmatrix} q_{comp,\mathrm{MP}},   \tag{14}\]
where $C = \frac{k_{s} L_{\mathrm{MP},\mathrm{IP}}}{EA R_\mathrm{IP}}$ is the compliance compensation coefficient for the thumb.

Since the CMC joint of the thumb is directly coupled to and controlled by the servo-6 output axis, the transmission chain is remarkably short and free of significant elastic load. Therefore, no mechanical compensation is required for this specific joint.

\begin{figure*}[t!]
    \centering
    \includegraphics[width=0.9\textwidth]{Pictures/table.jpg}
    \caption{\textbf{PHANTOM hand expressibility benchmark: hand gestures and grasping patterns.}
    Left: Hand gestures, including precise thumb-to-finger opposition (thumb opposing the index, middle, and ring fingers respectively), representative multi-finger combinations and digit gestures 0-8. Right: An extended set 16 representative grasps under the same controller, covering diverse contact topologies (enveloping, lateral, and tip pinches) and tool morphologies (rod-like, ring-like, and press-type), for cross-task consistency assessment.
    All results from the same control stack (analytic mapping + mechanics-based compensation), no additional sensing.}
    \label{fig:table}
\end{figure*}

\section{Experiments and Analysis}
\subsection{Device, Data Flow, and General Procedure}

\subsubsection{Hardware Platform}\

The experimental platform is the PHANTOM hand: a tendon-driven dexterous system with 6 actuators and 15 degrees of freedom (DoFs). It adopts an underactuated design with four fingers driven by a single servo via a figure-eight tendon routing, while the thumb employs two independent servos for the IP/MP and CMC joints. The hand is constructed at a 1:1 human scale and features magnet-based modular assembly. The joint indices ($q_{1}\!\sim q_{15}$) and actuator commands ($u_{1}\!\sim u_{6}$) are illustrated in Fig.~\ref{fig:phantom_overview}.

\subsubsection{Acquisition and Units}\

All angles are represented in radians unless otherwise specified. External measurements are ued exclusively for offline evaluation (i.e., not involved in closed-loop control). Each phalanx is augmented with a visible rigid reference (fiducial and edge model). Using a calibrated vision pipeline, we reconstruct the Euler angle of each joint (single-plane flexion–extension), denoted as $\mathbf{q_{\text{meas}}}$. To ensure consistency with the control layer, all recorded values are first converted to radians.
The sampled sequences are then processed with a second-order zero-phase Butterworth low-pass filter (cutoff frequency set according to motion bandwidth), followed by unwrapping and drift correction.

\subsubsection{Admissible Workspace Determination and Experimental Testing}\
During the admissible workspace determination phase, a single servo sweep is performed to determine the admissible workspace $[q_{\min}, q_{\max}]$. The workspace of each finger joint is detailed in Table \ref{tab:joint_workspace}, where the range is given in radians and represents the effective flexion-extension range of each joint under the designed control stack. In the testing phase, $\mathbf{H}$, $\mathbf{b(\cdot)}$, and the control gains are fixed. Performance is evaluated in three aspects: (i) static posture expressibility, (ii) grasp diversity, (iii) mapping accuracy via servo sweeps, and (iv) fingertip force evaluation. To verify reproducibility, the same protocol is repeated after cross-day reassembly (“re-test”), using the identical $\mathbf{H}$ and $\mathbf{b(\cdot)}$.   

\begin{table}[t]
\centering
\caption{\textbf{Admissible workspace of each finger joint of PHANTOM hand.} All values are in radians, corresponding to the joint indices $q_1 \sim q_{15}$.}
\label{tab:joint_workspace}
\renewcommand{\arraystretch}{1.2} 
\begin{tabular}{lcc}
\hline
Finger Joint & Joint Label & \makecell[c]{Admissible Workspace \\ $[q_{\min}, q_{\max}]$ (rad)} \\
\hline
Thumb IP     & $q_1$       & $[0.00, 0.99]$ \\
Thumb MP     & $q_2$       & $[0.00, 1.25]$ \\
Index DIP    & $q_3$       & $[0.00, 1.31]$ \\
Index PIP    & $q_4$       & $[0.00, 1.61]$ \\
Index MCP    & $q_5$       & $[0.00, 1.27]$ \\
Middle DIP   & $q_6$       & $[0.00, 1.28]$ \\
Middle PIP   & $q_7$       & $[0.00, 1.58]$ \\
Middle MCP   & $q_8$       & $[0.00, 1.24]$ \\
Ring DIP     & $q_9$       & $[0.00, 1.29]$ \\
Ring PIP     & $q_{10}$    & $[0.00, 1.59]$ \\
Ring MCP     & $q_{11}$    & $[0.00, 1.25]$ \\
Pinky DIP    & $q_{12}$    & $[0.00, 1.25]$ \\
Pinky PIP    & $q_{13}$    & $[0.00, 1.55]$ \\
Pinky MCP    & $q_{14}$    & $[0.00, 1.23]$ \\
Thumb CMC    & $q_{15}$       & $[0.00, 1.57]$ \\
\hline
\end{tabular}
\end{table}

\subsubsection{Control and Safety}\
The control input is computed online as 
\[
\mathbf{u^\star} = \mathbf{H^{+}\!\big(\tilde q_d + b(\tilde q_d)\big)}  \tag{15},
\]  
where $\mathbf{\tilde q_d}$ denotes the target posture clipped within $[\mathbf{q_{\min}}, \mathbf{q_{\max}}]$. A servo-space PD controller is applied, with velocity and torque saturation as well as range protection enforced at the driver level. An identical parameter set is maintained across all tasks to avoid tuning bias.

\subsection{Static Posture Expressibility}
This experiment evaluates the expressibility of static postures using 13 target configurations, as illustrated in Fig.~\ref{fig:table}(1-13). The set includes three categories: (i) opposition of the thumb to the index, middle, ring fingers (Fig.~\ref{fig:table}(1–3)); (ii) representative gestures made by multi-finger coordination (Fig.~\ref{fig:table}(4–7)); and (iii) digit gestures 1–8 (Fig.~\ref{fig:table}(8–13)).  

Each target posture $\mathbf{q_d}$ is derived from reference human hand geometry and mapped to the joint space following the indexing defined in Fig.\ref{fig:phantom_overview}.

\subsection{Grasp Diversity}
Grasp diversity is evaluated according to the established taxonomy of three primary grasp types: power, precision, and tool-specific. A representative set of grasping patterns is tested.
Each object category is evaluated independently in randomized order, with each object repeated for $N_{\text{trial}} = 30$ trials. The target grasp posture is first defined in joint space $\mathbf{q_d}$ (derived from reference human hand shapes), and then executed through the control input $\mathbf{u^\star}$. For grasps requiring a hand-shape transition (reach $\rightarrow$ closure $\rightarrow$ lift $\rightarrow$ place), all intermediate phases are performed under the same mapping and PD control scheme.  

Key observations focus on: (i) the ability of single-actuated fingers combined with the dual-actuated thumb to achieve opposition and lateral pinch; (ii) the ability to maintain consistent grasp stability across objects of varying diameters and weights; (iii) the posture-shaping capability in tool grasps, particularly the contribution of the CMC joint to lateral swing and opposition.  

The results (Fig. \ref{fig:table}(14-29)) show that power grasps achieve stable multi-finger enclosures for cylindrical, spherical, and stick-like tool handles. Precision grasps emphasize coordinated finger control in both force and positional accuracy, including three-finger and fingertip pinches for manipulating small objects such as keys and wires. Finally, tool-specific grasps correspond to functional postures for handling scissors, screwdrivers, tape rolls, tweezers, and other instruments.



\begin{figure*}[h]
    \centering
    \includegraphics[width=0.8\textwidth]{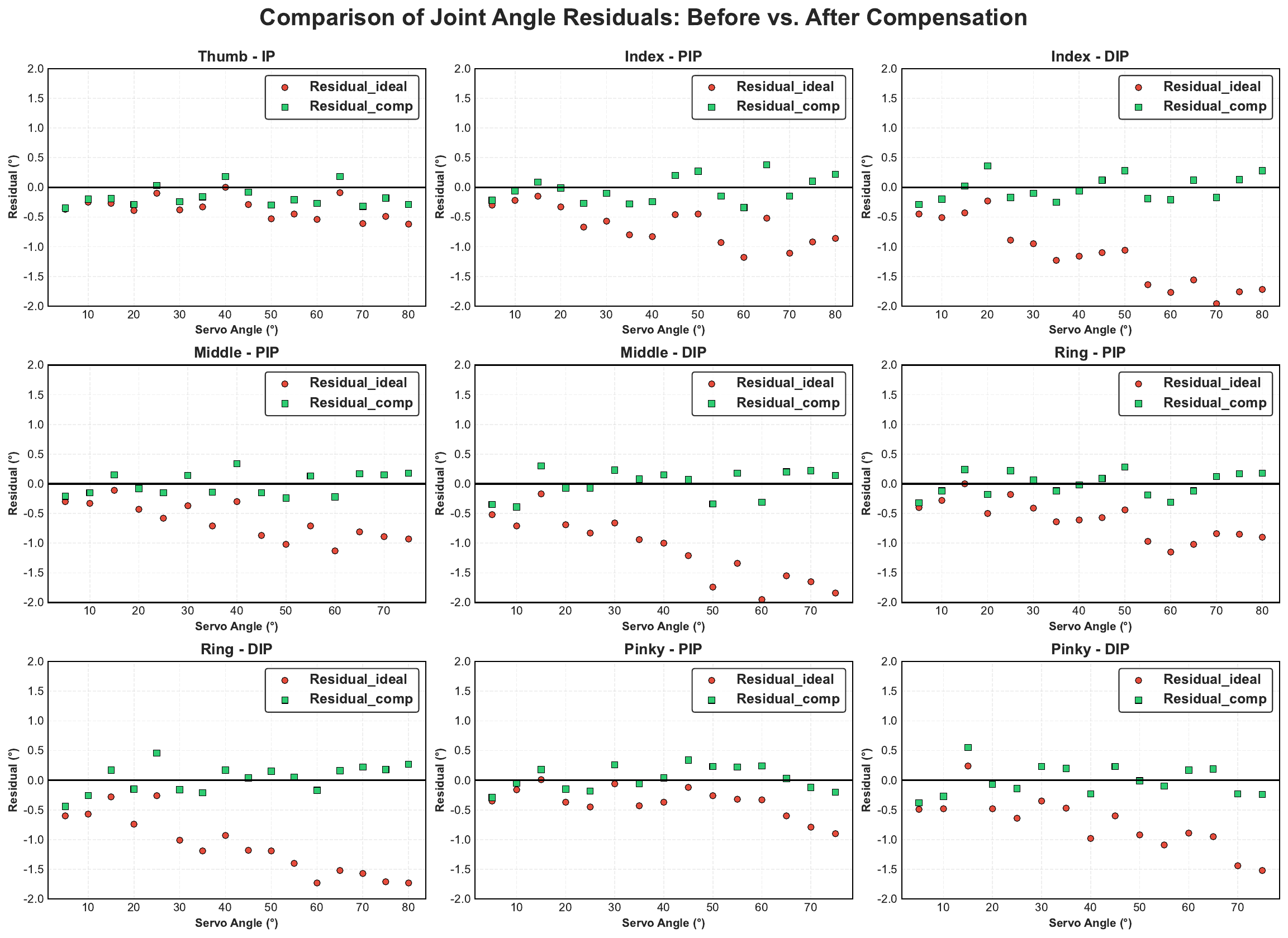}
    \caption{\textbf{Comparative analysis of joint angle residuals under uncompensated and compensated control.}
    This figure illustrates the relationship between the angle residuals and the servo control angles for target joints across the five fingers. The red circles represent the original uncompensated residuals (Residual\_{ideal}), while the green squares denote the residuals after mechanics-based compensation (Residual\_{comp}). All subplots are standardized with a Y-axis range from $-2.0$ to $+2.0$ degrees to facilitate a direct comparison of performance.}
    \label{fig:finger_residual_plot}
\end{figure*}

\subsection{Mapping Accuracy via Servo Sweeps}

\textbf{Objective.}  
This experiment quantifies the global accuracy and residual patterns of the ''analytic prior mapping + Mechanics-Based Compensation'' model, and compares it against the ``analytic-only'' baseline.  

\textbf{Method.}  
For each servo $j$, a monotonic sweep across its safe range is performed, discretized into $K_j$ sampling points. At each point, the corresponding joint angles $\mathbf{q_{\text{meas}}}$ are recorded. Using the analytic mapping model $\mathbf{q_{ideal}} = \mathbf{H(u)}\,\mathbf{u}$, the ideal joint values $\mathbf{q_{\text{ideal}}}$ are obtained. Using the compensation model $\mathbf{q_{comp}} = \mathbf{H(u)}\,\mathbf{u} - \mathbf{b(q)}$, the compensated joint angles $q_{\text{comp}}$ are obtained. 
Finally, two models are compared with respect to $\mathbf{q_{\text{meas}}}$. 

\textbf{Evaluation Metrics.}  
(i) Angular Residual: Defined as $\delta q = q_{\text{meas}} - q_{\text{model}}$, this metric captures the evolution of error patterns across the entire actuation range. Fig.~\ref{fig:finger_residual_plot} illustrates the residual distribution for 9 target joints (Thumb IP; PIP and DIP for others) across the full actuation range.

\begin{figure*}[h]
    \centering
    \includegraphics[width=0.8\textwidth]{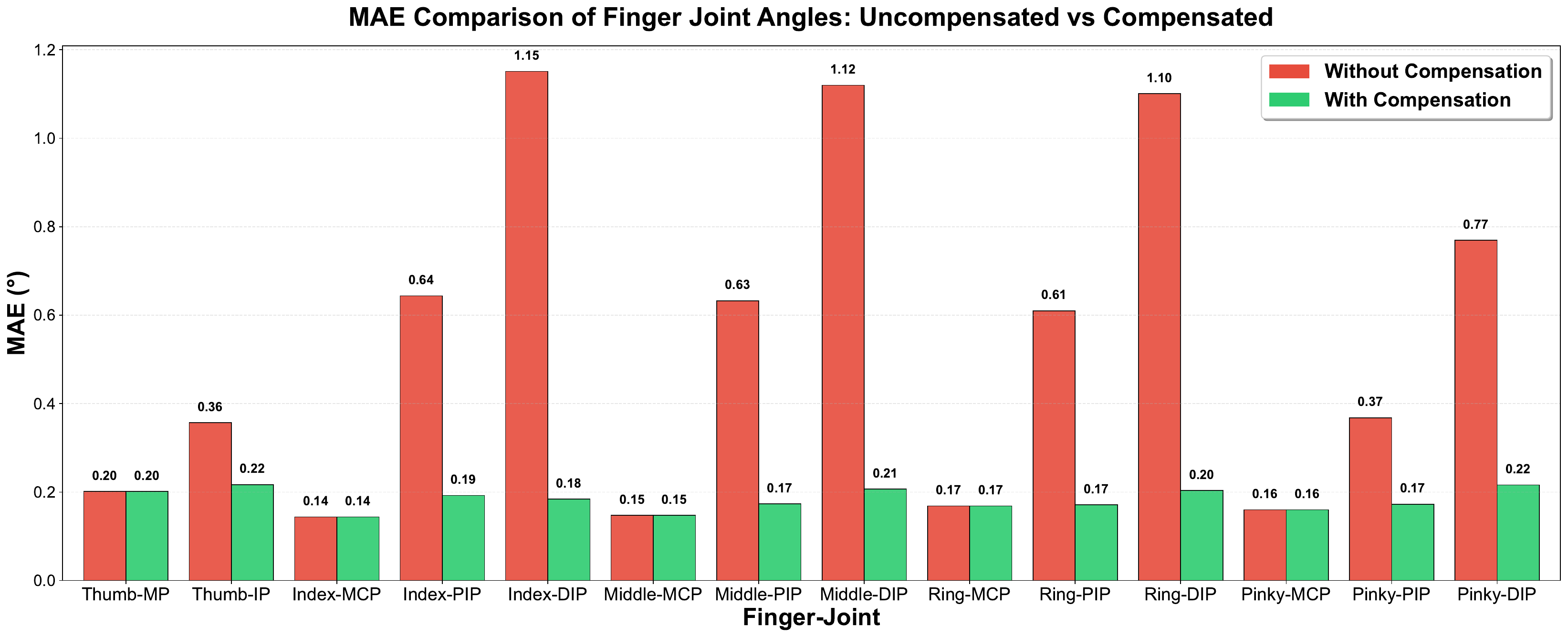}
    \caption{\textbf{Mean Absolute Error (MAE) comparison of joint angles for five fingers of PHANTOM Hand under uncompensated and compensated conditions.} The red bars represent the MAE between real joint angles and ideal joint angles (without compensation), while the green bars denote the MAE between real joint angles and compensation joint angles (with compensation). The fingers include the thumb (with MP and IP joints) and four fingers (each with MCP, PIP and DIP joints). The MAE is calculated in the unit of degrees ($^\circ$).}
    \label{fig:finger_mae_plot}
\end{figure*}

(ii) Mean Absolute Error (MAE): The MAE provides a statistical summary of the global prediction accuracy. The performance of both $\mathbf{q_{\text{ideal}}}$ and $\mathbf{q_{\text{comp}}}$ models is quantified and compared. The results are shown in Fig.~\ref{fig:finger_mae_plot}.


\textbf{Results and Analysis.} 
(i) As shown in Fig.~\ref{fig:finger_residual_plot}, the pre-compensation residuals exhibit distinct non-linear fluctuations and systemic offsets, with deviations approaching 2.0$^\circ$ in certain joints. After compensation, the residuals fluctuate closely around the $0^\circ$ baseline and maintain a uniform distribution across the entire motion range without significant trending bias. (ii) The Mechanics-Based Compensation model achieves a significant reduction in the MAE for the target joints (IP, PIP, DIP, excluding MCP and thumb MP). Among all target joints, the index finger DIP joint exhibits the most remarkable compensation effect, with its MAE decreasing sharply from 1.15 to 0.18. Meanwhile, the thumb IP joint and the PIP/DIP joints of other fingers also show obvious MAE reductions, which verifies the effectiveness of the proposed compensation model. (iii) After compensation, the MAE of all target joints is $\leq 0.23$, which falls within the sub-degree level. This indicates that the compensated joint angle predictions meet the high-precision requirements for practical application.

\begin{figure}[h]
    \centering
    \includegraphics[width=0.8\columnwidth]{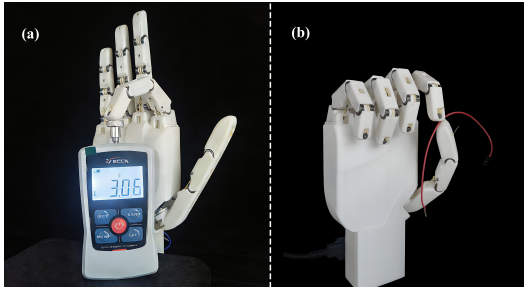}
    \caption{\textbf{Characterization of fingertip force and variable load manipulation of the PHANTOM hand.}
    (a) Quantitative measurement of fingertip output: the index finger exhibits a stable maximum force of 3.06 N. (b) Flexible wire clamping: the thumb and index finger cooperatively exert a minimal contact force to securely hold a thin wire.}
    \label{fig:force_evaluation}
\end{figure}

\subsection{Fingertip Force Evaluation}
\textbf{Objective.}
Fingertip force is a critical metric for assessing operational stability and payload capacity.
We conducted a quantitative characterization of fingertip output force for PHANTOM hand.

\textbf{Experimental Setup.}
The fingertip force was measured using a high-precision digital force gauge (as illustrated in Fig.~\ref{fig:force_evaluation}(a)). During the testing procedure, the actuators were commanded to execute a closing motion such that the fingertip exerted a normal force against the gauge's probe. To mitigate stochastic errors and ensure statistical significance, five independent trials were performed for each finger. The mean values were calculated and are summarized in Table \ref{tab:force_table}.

\textbf{Results and Analysis.}
(i) Table \ref{tab:force_table} shows that the average fingertip forces of all five digits are sufficient for stable grasping of common objects. Moreover, Fig. \ref{fig:force_evaluation}(b) illustrates the thumb and index finger securely holding a thin wire with minimal contact force, without causing structural deformation. These results demonstrate a wide dynamic force range, spanning from moderate-load grasping to low-load compliant manipulation. (ii) No significant elastic deformation was observed in the finger joints under peak force conditions, validating the mechanical rigidity and design optimization of the PHANTOM architecture.

\begin{table}
\centering
\caption{\textbf{Measured average fingertip forces of the PHANTOM hand.}
The data represent the mean output force of each finger calculated from five independent experimental trials using the digital force gauge shown in Fig.~\ref{fig:force_evaluation}.}
\label{tab:force_table}
\setlength{\tabcolsep}{10pt} 
\renewcommand{\arraystretch}{1.5} 
\begin{tabular}{lcccccc}
\hline
\textbf{Finger} & Thumb & Index & Middle & Ring & Pinky \\
\hline
\textbf{Force (N)} & 3.35 & 2.87 & 2.51 & 2.73 & 2.33 \\
\hline
\end{tabular}
\end{table}

\section{Conclusion}
This work addresses the long-standing challenge of achieving accurate, high-dimensional expressibility in underactuated systems by introducing the PHANTOM Hand: a 6-actuator, 15-DoF tendon-driven dexterous hand. To overcome the kinematic unpredictability inherent to compliant transmissions, a unified framework is established. It bridges an analytic geometric mapping with a mechanics-based compliance compensation model, naturally accounting for spring counter-tension and tendon elasticity. 

This hybrid design enables sub-degree reproducible posture expression (Fig.~\ref{fig:finger_mae_plot}) without the need for extensive additional sensorization. Static hand gestures and diverse grasping paradigms (Fig.~\ref{fig:table}) demonstrate the system's broad functional coverage. Crucially, quantitative fingertip force evaluations validate the core philosophy of PHANTOM: analytic precision is effectively exploited to govern complex kinematics, while the transmission retains robust, continuous force output to secure heavy tools. 

All resources—including hardware CAD models, URDF/ROS2 files, and control/calibration scripts—are open-sourced with this paper. The released content, as demonstrated in the accompanying video, provides a rigorous baseline to facilitate community collaboration and foster a thriving underactuated manipulation ecosystem.

\addtolength{\textheight}{-12cm}   



\bibliographystyle{IEEEtran} 
\bibliography{references}
\end{document}